# Hybrid Probabilistic Programs: Algorithms and Complexity

**Michael I. Dekhtyar**
Tver State University
Michael.Dekhtyar@tversu.ru

**Alex Dekhtyar**
University of Maryland
dekhtyar@cs.umd.edu

**V.S. Subrahmanian**
University of Maryland
vs@cs.umd.edu

## Abstract

Hybrid Probabilistic Programs (HPPs) are logic programs that allow the programmer to explicitly encode his knowledge of the dependencies between events being described in the program. In this paper, we classify HPPs into three classes called $HPP_1, HPP_2$ and $HPP_r, r \geq 3$. For these classes, we provide three types of results for HPPs. First, we develop algorithms to compute the set of all ground consequences of an HPP. Then we provide algorithms and complexity results for the problems of entailment ("Given an HPP $P$ and a query $Q$ as input, is $Q$ a logical consequence of $P$?") and consistency ("Given an HPP $P$ as input, is $P$ consistent?"). Our results provide a fine characterization of when polynomial algorithms exist for the above problems, and when these problems become intractable.

## 1 Introduction

Computing the probability of a complex event from the probability of the primitive events constituting it depends upon the dependencies (if any) known to exist between the events being composed. For example, consider two events $e_1, e_2$. The probability, $\mathbf{P}(e_1 \wedge e_2)$ of the occurrence of both is events is 0 if the events are mutually exclusive. However, if the events are independent, then $\mathbf{P}(e_1 \wedge e_2) = \mathbf{P}(e_1) \times \mathbf{P}(e_2)$. If we are ignorant of the relationship between these two events, then, as stated by Boole[1], the best we can say about $\mathbf{P}(e_1 \wedge e_2)$ is that it lies in the interval $[\max(0, \mathbf{P}(e_1) + \mathbf{P}(e_2) - 1), \min(\mathbf{P}(e_1), \mathbf{P}(e_2)]$.

In short, computing the probability of a complex event depends fundamentally upon our knowledge about the dependences between the events involved. In [2] we proposed a language called *Hybrid Probabilistic (Logic) Programs* (or HPPs, for short), that extended logic programs to deal with diverse types of probabilistic dependencies, and we defined the semantics of such a language. HPPs build upon the idea of an annotated logic program introduced in [21], and studied extensively by many researchers over the years [6, 9, 8] In this work, we make two classes of contributions.

First, we study the complexity of a variety of problems related to the semantics of HPPs. In particular, we show that the complexity of the entailment problem (answers to queries to HPPs) is polynomial for HPPs with atomic heads of rules, and in many cases for HPPs with at most two atoms in the heads. However, when formulas of size three or more are allowed in the heads of the rules, the complexity of query processing becomes NP-complete. We establish some other complexity results for related problems, such as checking the consistency of an HPP.

Second, we propose a proof system $HGR_P$ for HPPs that may be used for query processing. This is a Hilbert-style proof system and it is shown to be sound and complete. We show that proofs in $HGR_P$ are polynomially bounded in size (this is consistent with the preceding NP-completeness result because the search space may involve exponentially many derivations each of polynomially bounded length). This is an interesting and counterintuitive result — it says that (the answers to) all queries to HPPs have at least one polynomial explanation. It is well-known (see e.g. [20]) that for propositional classical logic, an existence of proof systems with polynomially bounded length of proofs is a difficult open question, as an affirmative answer implies that $NP = coNP$. In fact, for many proof systems for classic propositional logic (e.g. resolution based) and for variety of nonmonotonic logics superpolynomial lower bounds were established ([23, 20]).

Section 2 recapitulates the syntax and semantics of HPPs as described in [2]. In Section 3, we describe results on the computation complexity of HPPs. Section 4 introduces the proof system $HGR_P$, shows it is sound and complete, and then presents results showing the proofs in $HGR_P$ are polynomially bounded.

## 2 Background

The aim of this section is to describe the syntax and semantics of HPPs — the content of this section is not new and overviews results in [2]. HPPs are based on an abstract class of functions called *probabilistic strategies*. Associated with each such strategy $s$, we can introduce a new "conjunction like" connective $\wedge_s$ and a new "disjunction like" connective, $\vee_s$, which may then be used to define a syntax for HPPs.

### 2.1 Probabilistic Strategies (p-strategies)

It is well-known that the probability of a compound event may be an *interval*, rather than a point even if point probabilities are known for the primitive events involved. This was first shown by Boole[1] in 1854. Thus, p-strategies will be defined on intervals – points, in any case, are special cases of intervals.

**Definition 1** *A probabilistic strategy (p-strategy) is a pair of functions:* $\rho = \langle c, md \rangle$, *such that:*
1. $c : \mathcal{C}[0,1] \times \mathcal{C}[0,1] \longrightarrow \mathcal{C}[0,1]$ *is called a probabilistic composition function if it satisfies the following axioms:*
*(a)* Commutativity :
$c([a_1, b_1], [a_2, b_2]) = c([a_2, b_2], [a_1, b_1])$
*(b)* Associativity :
$c(c([a_1, b_1], [a_2, b_2]), [a_3, b_3]) = c([a_1, b_1], c([a_2, b_2], [a_3, b_3]))$
*(c)* Monotonicity :
$c([a_1, b_1], [a_2, b_2]) \subseteq c([a_3, b_3], [a_2, b_2])$ if $[a_1, b_1] \subseteq [a_3, b_3]$
*(d)* Separation: *there exist two functions $c^1$ and $c^2$ such that* $c([a_1, b_1], [a_2, b_2]) = (c^1(a_1, a_2), c^2(b_1, b_2))$
2. $md : \mathcal{C}[0,1] \longrightarrow \mathcal{C}[0,1]$ *is called a maximal interval function.*

Intuitively, a composition function determines, given the probability ranges of two events, the probability range of their (either and- or or-composition). A max-interval function $md$ returns the best estimate for the probability of simple event given the probability of a compound event. For the discussion on why we specify max-interval functions as above see [2].

The two combinations of events we plan on dealing with are conjunctions of events and disjunction of events. Among all possible p-strategies, we identify *conjunctive* and *disjunctive* p-strategies, which will handle the computation of probabilities of these two combinations respectively.

Since composition functions are both commutative and associative, all terms constructed by applications of composition function $c$ to $n \geq 2$ intervals $\mu_1 = [a_1, b_1], \ldots, \mu_n = [a_n, b_n]$ will have the same value which we will denote as $c(\mu_1, \ldots, \mu_n)$ with it's lower bound $c^1(a_1, \ldots, a_n)$ and upper bound $c^2(b_1, \ldots, b_n)$. For technical reasons it's convenient in the case $n = 1$ for any $\mu = [a, b]$ to set $c(\mu) = \mu, c^1(a) = a$ and $c^2(b) = b$.

In the definition below $[a, b] \leq [c, d]$ means that $a \leq b$ and $c \leq d$.

**Definition 2** Conjunctive and Disjunctive p-strategies
A *p-strategy* $< c, d >$ *is called* conjunctive (disjunctive) *if it satisfies the following axioms:*

| Axiom | Conjunctive Strategies |
|---|---|
| Bottomline | $c([a_1, b_1], [a_2, b_2]) \leq [min(a_1, a_2), min(b_1, b_2)]$ |
| Identity | $c([a, b], [1, 1]) = [a, b]$ |
| Annihilator | $c([a, b], [0, 0]) = [0, 0]$ |
| Max.Interval | $md([a, b]) = [a, 1]$ |
| Axiom | Disjunctive Strategies |
| Bottomline | $[max(a_1, a_2), max(b_1, b_2)] \leq c([a_1, b_1], [a_2, b_2])$ |
| Identity | $c([a, b], [0, 0]) = [a, b]$ |
| Annihilator | $c([a, b], [1, 1]) = [1, 1]$ |
| Max.Interval | $md([a, b]) = [0, b]$ |

For a more complete discussion of the axioms we refer the reader to [2].

**Example 1** *Below are some examples of p-strategies. We provide definitions of composition functions only, as max-interval functions are defined uniquely by the type of p-strategy [2].*

- *inc:* p-strategies for independence assumption
  *Conjunctive:* $c_{inc}([a, b], [c, d]) = [ac, bd]$.
  *Disjunctive:*
  $c_{ind}([a, b], [c, d]) = [a + c - ac, b + d - bd]$.

- *igc:* p-strategies for ignorance assumption
  *Conjunctive:*
  $c_{igc}([a, b], [c, d]) = [max(0, a + c - 1), min(b, d)]$.
  *Disjunctive:*
  $c_{igd}([a, b], [c, d]) = [max(a, c), min(1, b + d)]$.

- *pcc:* p-strategies for positive correlation assumption
  *Conjunctive:*
  $c_{pcc}([a, b], [c, d]) = [min(a, c), min(b, d)]$.
  *Disjunctive:*
  $c_{pcd}([a, b], [c, d]) = [max(a, c), max(b, d)]$.

- p-strategy for negative correlation assumption
  *Disjunctive:*
  $c_{ncd}([a, b], [c, d]) = [min(a + c, 1), min(b + d, 1)]$

**Example 2** *We illustrate why max-interval function as defined as above on the following example. Conider conjunctive p-strategy for independence inc. Suppose* we know that the probability of the conjunction of two events $e_1$ and $e_2$ under the assumption of independence lies in the interval $[a, b]$. Our goal is to find the interval in which the probability of each of the two simple events lies. We are looking at pairs of intervals $[a_1, b_1]$ and $[a_2, b_2]$ such that $a_1 a_2 = a$ and $b_1 b_2 = b$. Clearly we are interested in minimal possible values for $a_1$ and $a_2$ and maximal possible values of $b_1$ and $b_2$ such that the above equalities hold. It is easy to notice that both $a_1$ and $a_2$ can go as low as $a$ while both $b_1$ and $b_2$ can reach 1 (not together). Just set $[a_1, b_1] = [a, b]$ and $[a_2, b_2] = [1, 1]$ or vice versa. So, 1 is the maximum number $b_1$ and $b_2$ can reach. Now we note that in order for a product of two numbers less than or equal to 1 to be equal to a number $a$, neither number can be less than $a$. This makes $a$ the minimum $a_1$ and $a_2$ can reach. This suggests that the probability that each of the events $e_1$ and $e_2$ holds lies between $a$ and 1. This interval is what will be returned by the $md_{inc}$ function.

As this paper investigates complexity of some algorithmic problems related to HPPs, we assume that all intervals are bounded by rational numbers (which may be represented for example by finite binary numbers). To make our results independent of complexity of particular strategies we will assume below that the computation of a composition function for each p-strategy is provided by a constant time oracle. This way, all bounds obtained in this paper should be multiplied by the complexity of computing the composition. However, for composition functions computable in polynomial time such multiplication will not result in the change in the polynomiality (deterministic or nondeterministic) of the bounds.

## 2.2 Syntax of hp-programs

Let $L$ be a language which has predicate, variable and constant symbols, but has no function symbols. Let $B_L$ be the set of all ground atoms of $L$.

In hybrid probabilistic programs, we assume the existence of an arbitrary, but fixed set of conjunctive and disjunctive p-strategies $S$ denote $CONJ \cup DISJ$. The programmer may augment this set with new strategies when s/he needs new ones for their application. Each conjunctive p-strategy $\rho$ has an associated conjunction operator $\wedge_\rho$ and each disjunctive p-strategy $\rho'$ has an associated disjunction operator $\vee_{\rho'}$.

Hybrid basic formulas, defined below, are either conjunctions of atoms, or disjunctions of atoms (but not mixes of both) w.r.t. a single connective.

**Definition 3** *Let $\rho$ be a conjunctive p-strategy, $\rho'$ be a disjunctive p-strategy and $A_1, \ldots, A_k$ be distinct atoms. Then $A_1 \wedge_\rho \ldots \wedge_\rho A_k$ and $A_1 \vee_{\rho'} A_2 \ldots \vee_{\rho'} A_k$ are called* hybrid basic formulas. *Suppose $bf_\rho(B_L)$ denotes the set of all ground hybrid basic formulas for the $\vee_\rho$ and $\wedge_\rho$ connectives. Let $bf_S(B_L) = \cup_{\rho \in S} bf_\rho(B_L)$.*

We define a notion of *annotations* inductively as follows: (1) Any real number or variable over real numbers is an annotation term. (2) If $f$ is an interpreted function over the reals of arity $k$ and $t_1, \ldots, t_k$ are annotation terms, then $f(t_1, \ldots, t_k)$ is an annotation term. An *annotation* is pair $[at_1, at_2]$ where $at_1, at_2$ are annotation terms. Thus, for instance, $[0.5, 0.6]$, $[0..5, \frac{V+1}{2}]$ are both annotations.

**Definition 4** *A hybrid probabilistic annotated basic formula (hp-annotated basic formula) is an expression of the form $B : \mu$ where $B$ is a hybrid basic formula and $\mu$ is an annotation.*

**Definition 5** *Let $B_0, B_1, \ldots, B_k$ be hybrid basic formulas. Let $\mu_0, \mu_1, \ldots, \mu_k$ be annotations. A hybrid probabilistic clause (hp-clause) is a construction of the form: $B_0 : \mu_0 \leftarrow B_1 : \mu_1 \wedge \ldots \wedge B_k : \mu_k$.*

Informally speaking, the above rule is read: "If the probability of $B_1$ falls in the interval $\mu_1$ and $\cdots$ the probability of $B_k$ falls within the interval $\mu_k$, then the probability of $B_0$ falls within the interval $\mu_0$. Note that it is entirely possible that $B_i$ uses a connective $\wedge_\rho$ corresponding to a particular (conjunctive) p-strategy, while $B_j$ may use a connective $\vee_{\rho'}$ corresponding to some other disjunctive p-strategy. HPPs allow mixing and matching of different kinds of p-strategies, both in the $B_i$'s in the body of a rule, as well as in $B_0$ - the head of a rule.

**Definition 6** *A hybrid probabilistic program (hp-program) over set $S$ of p-strategies is a finite set of hp-clauses involving only connectives from $S$.*

An hp-program is *ground* iff its every clause is ground, i.e. all its clauses do not contain neither variables nor variable annotations.

For example, consider an image processing application that contains a set of facts stating who was seen with whom. These facts may be extracted by an image processing program which may identify persons in images with associated probabilities. A higher level program then classifies individuals as suspects based on different criteria. Such an application may be encoded as an hp-program containing rules such as those shown below.

seen(pic1,id1,john) : [0.5, 0.7] ←
seen(pic1,id1,ed) : [0.2, 0.4] ←
seen(pic1,id2,ed) : [0.5, 0.6] ←
seen(pic1,id2,dan) : [0.2, 0.5] ←
suspect1($X$) : [1, 1] ←
       seen($Pic, Id1, X$) : [0.5, 1] ∧
       seen($Pic, Id2,$ ed) : [0.5, 1] ∧ $Id1 \neq Id2$.

suspect2($X$): [1, 1] ←
  (seen($Pic,Id1,X$) $\wedge_{ig}$ seen($Pic,Id2$,ed)): [0.5, 1].
suspect3($X$): [1, 1] ←
  (seen($Pic,Id1,X$) $\wedge_{in}$ seen($Pic,Id2$,ed)): [0.5, 1].

In the above example, we have two pictures, each of which contains two objects. Picture $pic1$'s object with $id2$ is identified as Ed with 50-60% probability and Dan with 20-50% probability. Three alternative definitions of suspect are given. The first says that if X occurs (with over 50% probability) in a picture where Ed also appears (with over 50% probability), then X is considered a suspect. By this rule, John is a suspect. The second rule says that if we know nothing about the occurrences of people in a picture and if the probability that Ed and X are both in the picture is over 50% under this assumption, then X is considered a suspect. According to this rule, there are no suspects at all. A third possibility is that X be considered a suspect if we assume that people's appearances in pictures are independent of one another, and under this assumption, Ed and X are both in the picture with over 50% probability. This rule yields no suspects either.

### 2.3 Fixpoint and Model Theory for hp-programs

In this subsection, we briefly describe the model theory underlying HPPs. [2] contains a more comprehensive description. Before proceeding further we first introduce some notation for "splitting" a complex formula into two parts.

**Definition 7** Let $F = F_1 *_\rho \ldots *_\rho F_n$, $G = G_1 *_\rho \ldots *_\rho G_k$ $(k > 0)$, $H = H_1 *_\rho \ldots *_\rho H_m$ $(m > 0)$ where $* \in \{\wedge, \vee\}$. We will write $G \oplus_\rho H = F$ (or $G \oplus H$ if the p-strategy $\rho$ is irrelevant) iff:
(a) $\{G_1, \ldots, G_k\} \cup \{H_1, \ldots, H_m\} = \{F_1, \ldots, F_n\}$ and
(b) $\{G_1, \ldots, G_k\} \cap \{H_1, \ldots, H_m\} = \emptyset$.

The analog of an Herbrand interpretation in classical logic programs is what we call a *hybrid formula function*.

**Definition 8** *A function* $h : bf_S(B_L) \longrightarrow \mathcal{C}[0, 1]$, *is called a* hybrid formula function *iff it satisfies the following three conditions:*
1. Commutativity. If $F = G_1 \oplus_\rho G_2$ then $h(F) = h(G_1 *_\rho G_2)$.
2. Composition. If $F = G_1 \oplus_\rho G_2$ then $h(F) \subseteq c_\rho(h(G_1), h(G_2))$.
3. Decomposition. For any basic formula $F$, $h(F) \subseteq md_\rho(h(F *_\rho G))$ for all $\rho \in S$ and $G \in bf_S(B_L)$.

From the first condition it follows that $h(F) = h(F')$ for any $F$ and $F'$ which are permutations of one another. Second condition states that the probability of a complex formula is bounded by the probabilities of its subformulas. Conversely, the third condition bounds the probability of a subformula by the probability of a formula it is a part of. We say that hybrid formula function $g$ is less than or equal to hybrid formula function $h$, denoted $g \leq h$ iff $(\forall F \in bf_S(B_L))(g(F) \supseteq h(F))$.

We are now in a position to specify what it means for a hybrid basic formula function to satisfy a formula.

**Definition 9** Satisfaction. Let $h$ be a hybrid basic formula function, $F \in bf_S(B_L)$, $\mu \in \mathcal{C}[0, 1]$. We say that

- $h \models F : \mu$ iff $h(F) \subseteq \mu$.

- $h \models F_1 : \mu_1 \wedge \ldots \wedge F_n : \mu_n$ iff $(\forall 1 \leq j \leq n) h \models F_j : \mu_j$.

- $h \models F : \mu \longleftarrow F_1 : \mu_1 \wedge \ldots \wedge F_n : \mu_n$ iff either $h \models F : \mu$ or $h \not\models F_1 : \mu_1 \wedge \ldots \wedge F_n : \mu_n$.

- $h \models (\exists x)(F : \mu)$ iff $h \models F(t/x) : \mu$ for some ground term $t$.

- $h \models (\forall x)(F : \mu)$ iff $h \models F(t/x) : \mu$ for every ground term $t$.

*A formula function $h$ is called a* model *of an hp-program $P$ ($h \models P$) iff ($h \models C$) for all clauses $C \in P$.*

As usual, we say that $F : \mu$ is a *consequence* of $P$ iff for every model $h$ of $P$, it is the case that $h(F) \subseteq \mu$.

It is possible for a hybrid formula function $h$ to assign $\emptyset$ to some formula. When $h(F) = \emptyset$, $h$ is "saying" that $F$'s probability lies in the empty set. This corresponds to an inconsistency because, by definition, nothing is in the empty set.

**Definition 10** *Formula function $h$ is called* fully defined *iff* $\forall (F \in bf_S(B_L))(h(F) \neq \emptyset)$.

Now we introduce the fixpoint semantics for the hp-programs. Operator $S_P$ is a preliminary operator, restricted only to the clauses which have the same head as the argument. It is then extended to full fixpoint operator $T_P$

**Definition 11** *Let $P$ be a hybrid probabilistic program. Operator $S_P : \mathcal{HFF} \longrightarrow \mathcal{HFF}$ is defined as follows (where $F$ is a basic formula):* $S_P(h)(F) = \cap M$ *where* $M = \{\mu\sigma | F : \mu \longleftarrow F_1 : \mu_1 \wedge \ldots \wedge F_n : \mu_n$ *is a ground instance of some clause in $P$ ; $\sigma$ is a ground substitution of annotation variables and $(\forall j \leq n) h(F_j) \subseteq \mu_j \sigma\}$ if $M = \emptyset$ $S_P(h)(F) = [0, 1]$.*

We use the definition of $S_P$ to define the immediate consequence operator $T_P$.

**Definition 12** Let $P$ be a hybrid probabilistic program. We inductively define operator $T_P : \mathcal{HFF} \longrightarrow \mathcal{HFF}$ as follows:

1. Let $F$ be an <u>atomic</u> formula.
 * if $S_P(h)(F) = \emptyset$ then $T_P(h)(F) = \emptyset$.
 * if $S_P(h)(F) \neq \emptyset$, then let $M = \{\langle\mu\sigma, \rho\rangle | (F \oplus_\rho G) : \mu \longleftarrow F_1 : \mu_1 \wedge \ldots \wedge F_n : \mu_n \text{ where } \sigma$ is a ground substitution for the annotation varables and $i \in S$ and $(\forall j \leq n) h(F_j) \subseteq \mu_j\sigma\}$. We define
$T_P(h)(F) = (\cap\{md_\rho(\mu\sigma) | \langle\mu\sigma, i\rangle \in M\}) \cap S_P(h)(F)$.

2. ($F$ <u>not atomic</u>) Let $F = F_1 *_\rho \ldots *_\rho F_n$.
Let $M' = \{\langle\mu\sigma, \rho\rangle | D_1 *_\rho \ldots *_\rho D_k : \mu \longleftarrow E_1 : \mu_1 \wedge \ldots E_m : \mu_m \in ground(P);$
$(\forall 1 \leq j \leq m) h(E_j) \subseteq \mu_j; \{F_1, \ldots, F_n\} \subset \{D_1, \ldots D_k\}, n < k\}$ Then:

$T_P(h)(F) = S_P(h)(F) \cap (\cap\{c_\rho(T_P(h)(G), T_P(h)(H)) |$
$G \oplus_\rho H = F\}) \cap (\cap\{md_\rho(\mu\sigma) | \langle\mu\sigma, i\rangle \in M'\})$

In [2] it was shown that both $S_P$ and $T_P$ are monotonic if the annotations of the atoms in $P$ are constant.

**Definition 13** 1. $T_P^0 = h_\perp$ where $\perp$ is the atomic function that assigns $[0, 1]$ to all ground atoms $A$.
2. $T_P^\alpha = T_P(T_P^{\alpha-1})$ where $\alpha$ is a successor ordinal whose predecessor is denoted by $\alpha - 1$.
3. $T_P^\gamma = \sqcup\{T_P^\alpha | \alpha < \gamma\}$, where $\gamma$ is limit ordinal.

The following results [2] ties together, the fixpoint theory and the model theoretical characterizations of hp-programs, regardless of which p-strategies occur in the hp-program being considered.

**Theorem 1** Let $P$ be any hp-program. Then: 1. $h$ is a model of $P$ iff $T_P(h) \leq h$.
2. $P$ has a model iff $lfp(T_P)$ is fully defined.
3. If $lfp(T_P)$ is fully defined, then it is the least model of $P$, and $F : \mu$ is a logical consequence of $P$ iff $lfp(T_P) \subseteq \mu$.

In what follows we will consider only ground hp-programs. It is clear that for any such program $P$, the least fixpoint of its $T_P$ operator, $lfp(T_P)$ is achieved in a finite number of iterations, i.e., at least, $lfp(T_P) = T_P^\omega$. For brevity we will denote $lfp(T_P)$ as $h_P$.

## 3 Algorithms and Complexity of ground HPPs

In this section, we will develop algorithms, and associated complexity results, for three kinds of HPP problems: logical consequences of an HPP $P$, entailment problem (answer to query), and consistency of $P$. Obviously, these three problems are closely related to one another. Due to space restrictions we are able to present only the algorithms and state the theorems here. All proofs and reductions can be found in [3].

### 3.1 Complexity of model computation

Given a basic formula $G$, we define the $width(G)$ to be the number of atoms in $G$.
Given an hp-clause $C = B_0 : \mu_0 \leftarrow B_1 : \mu_1 \wedge \ldots \wedge B_k : \mu_k$, we say that the *head-width* of $C$ is the width of $B_0$, and the *body-width* of $C$ is $\max\{width(B_1), \ldots, width(B_k)\}$.

We may now define a hierarchy of subclasses of HPPs in terms of the head/body widths of the clauses involved.

**Definition 14** Let $HPP_{k,r}$ denote the class of HPP-programs $P$ such that for all clauses $C \in P$, the head-width of $C$ is less than or equal to $k$ and the body-width of $C$ is less than or equal to $r$. Let $HPP_k = \bigcup_{r \geq 0} HPP_{k,r}$.

Algorithm **LFP** below shows how we may compute the least fixpoint of $T_P$ for class $HPP_{k,r}$.

**Algorithm LFP.**
Input: $P \in HPP_{k,r}$, $N \geq \max(k, r)$, $m$ - number of clauses in $P$, $F_1, \ldots F_M$ - all formulas of $width \leq N$, lexicographically ordered.
Output: table $t_{2m}(F_k), 1 \leq k \leq M$.
BEGIN (algorithm)
(1) FOR $j = 1$ TO $M$ DO $t_0(F_j) := [0, 1]$;
(2) FOR $i = 0$ TO $2m - 1$ DO
 BEGIN
(3)  FOR EACH $C = G : \mu \leftarrow G_1 : \mu_1 \wedge \ldots \wedge G_n : \mu_l \in P$ such that for all $1 \leq j \leq n$ $t_i(G_j) \subseteq \mu_j$ DO
 BEGIN
(4)   $t_{i+1}(G) := t_i(G) \cap \mu$;
(5)   delete $C$ from $P$;
(6)   FOR EACH $H$ included in $G$, i.e. $G = H \oplus_\rho H'$ DO
(7)    $t_{i+1}(H) := t_i(H) \cap md_\rho(\mu)$;
 END
(8)  FOR $j = 1$ TO $M$ DO
(9)   FOR $k = j + 1$ TO $M$ DO
(10)    IF $F_k = F_j \oplus_\rho F_l$ (for some $l < k$)
(11)    THEN $t_{i+1}(F_k) := t_{i+1}(F_k) \cap c_i(t_{i+1}(F_j), t_{i+1}(F_l))$;
 END
END.

The following theorem proves that algorithm **LFP** is a correct way of computing the least fixpoint of $T_P$ for class $HPP_{k,r}$ and establishes its complexity.

**Theorem 2** Let $P$ be any program in $HPP_{k,r}$ with $m$ clauses. Let $a$ be the number of different atoms in $P$, $s$ be the number of different strategies in $P$, and $N \geq max\{k, r\}$. Then **Algorithm LFP** *computes $h_P$ on all the formulas of $bf_S(B_L)$ of width $\leq N$ in time $O(2m(2sa^N)^2) = O(m(sa^N)^2)$.*

The proof of this result is long and complex, and uses the properties that (i) if a program $P$ consists of $m$ ground clauses, then $T_P^{2m} = lfp(T_P)$ and that (ii) for every $i = 0, 1, \ldots, 2m$ and for every $k = 1, \ldots M$, the quantity $t_i(F_k)$ in algorithm **LFP** coincides with $T_P^i(F_k)$ and therefore $t_{2m}(F_k) = h_P(F_k)$.

It is important to note that this theorem tells us that computing the least fixpoint of an HPP is *exponential* in the width of the largest formula of interest. In other words, if we were to develop an implementation of HPPs, and we required that no basic formulas of length greater than $\delta$ for some fixed $\delta$ are allowed, then the above theorem yields a polynomial result. This is a reasonable assumption, as we do not expect that formulas of width greater than some small constant (e.g., 4) would be of interest in any practical application. This is stated in the following corollary.

**Corollary 1** *Let $P$ be any hp-program, and suppose $\delta$ is a fixed bound on the width of basic formulas occurring in $P$. Then $h_P$ can be computed in polynomial time of size of $P$ for all formulas of width $\leq \delta$.*

## 3.2 Complexity of Entailment

While it is important to know the complexity of computing the entire model of an HPP, it is really the entailment problem which gets solved over and over when queries are asked to the program. In this section we will consider the complexity of entailment problem on HPPs: given a *consistent* program $P$ and a query $F : \mu$, check whether $P \models F : \mu$.

As usual we fix some standard encoding which is used to represent programs and queries. If $P$ is an HPP, $|P|$ will denote the size of the representation of $P$ in this encoding, and similarly, if $F : \mu$ is an annotated basic formula, $|F : \mu|$ will denote the size of its representation. The complexity results in the sections that follow will be relative to $|P|$ and $|F : \mu|$.

In order to carry out our analysis of the entailment problem, we will split the results into three parts based on the syntax of HPPs.

At first, we show that if we consider the class $HPP_1$ containing only atoms in rule heads, then we can specialize algorithm **LFP** to a better algorithm, **LFP**$_1$ for computing the least fixpoint of $T_P$.

**Algorithm LFP**$_1$.
Input: $P \in HPP_1$, $m$ - number of clauses in $P$, $F_1, \ldots, F_M$ - lexicographical enumeration of all formulas in $P$, $F = A_1 *_\rho \ldots *_\rho A_n$, $A_i \in B_L, i \in \{1, \ldots, n\}$.
Output:  $\mu'$ - a subinterval of $[0, 1]$

BEGIN
(1)    FOR $j = 1$ TO $M$ DO $t(F_j) := [0; 1]$;
(2)    FOR $i = 0$ TO $2m - 1$ DO
       BEGIN
(3)        FOR EVERY $C = G : \mu \leftarrow G_1 : \mu_1, \ldots, G_l : \mu_l \in P$ such that $t(G_j) \subseteq \mu_j$ for all $1 \leq j \leq l$ DO
           BEGIN
(4)            $t(G) := t(G) \cap \mu$;
(5)            delete $C$ from $P$
           END
(6)        FOR $k = 1$ TO $M$ DO
(7)        IF $F_k = B_1 *_\rho \ldots *_\rho B_r$, $r > 0$, $B_j$ is an atom for all $1 \leq j \leq r$
(8)            THEN $t(F_k) := c_\rho(t(B_1), \ldots, t(B_r))$;
       END;
(9)    $\mu' := c_\rho(t(A_1), \ldots, t(A_n))$
END. (algorithm)

The following result specifies that the above algorithm may be directly used to check if an annotated basic formula $F : \mu$ is entailed by an HPP $P \in HPP_1$. Find the value $\mu'$ returned by Algorithm **LFP**$_1$ — if $\mu' \subseteq \mu$, then $P \models F : \mu$, else it does not.

**Theorem 3** *For any program $P \in HPP_1$ and annotated basic formula $F : \mu$ the entailment problem "$P \models F : \mu$" is solvable in time $O(|P|^2 + |F : \mu|)$ via Algorithm **LFP**$_1$.*

Our next goal is to develop algorithms and provide complexity results for checking entailment when we consider hybrid probabilistic programs for $HPP_r$ when $r \geq 3$.

We start our analysis by first considering the class $HPP^0$ of HPPs that only consist of facts, i.e. all rules in such HPPs have an empty body.

The following *nondeterministic* algorithm allows check entailment for hybrid probabilistic programs in this class.

**Algorithm Ent-HPP**$^0$
Input: program $P \in HPP^0$, consisting of $m$ clauses $C_1, \ldots, C_m$ with empty bodies: $C_k = H_k : [a_k, b_k]$, $k = 1, \ldots, m$, formula $F = B_1 *_\rho \ldots *_\rho B_n$, and an interval $[a, b]$.
1. Guess such $k \leq width(F)$, sequence of bounds $x_1, \ldots, x_k$ and partition $F = F_1 *_\rho \ldots *_\rho F_k$ which satisfy conditions:
(i) each $F_i$ is either some head $H_k$ and $x_i \geq b_k$, or it is a subformula of some head $H_k = F_i \oplus_\rho H'_k$ and $md_\rho([a_k, b_k]) \subseteq [0, x_i]$, and
(ii) $c_\rho^2(x_1, \ldots, x_k) \leq b$.
2. Guess such $k \leq width(F)$, sequence of bounds $x_1, \ldots, x_k$ and partition $F = F_1 *_\rho \ldots *_\rho F_k$ which satisfy conditions:
(i) each $F_i$ is either some head $H_k$ and $x_i \leq a_k$, or a subformula of some head $H_k = F_i \oplus_\rho H'_k$ and $md_\rho([a_k, b_k]) \subseteq [x_i, 1]$, and
(ii) $c_\rho^1(x_1, \ldots, x_k) \geq a$.
3. If both attempts are successful then output "Yes".

The following lemma establishes correctness of Algorithm **Ent-HPP$^0$** and its complexity.

**Lemma 1** *(1) For any consistent hp-program $P \in HPP^0$ and any query $F : [a,b]$ algorithm Ent-HPP$^0$ output "Yes" iff $P \models F : [a,b]$.*
*(2) Algorithm Ent-HPP$^0$ works in nondeterministic polynomial time.*

Let us denote by $Fired_P(i)$ the set of those clauses of $P$ whose bodies are satisfied by $T_P^i$ but are not satisfied by $T_P^{i-1}$.

We are now ready to present a generic algorithm, Algorithm **Ent-HPP**, that computes entailment by HPPs.

**Algorithm Ent-HPP**
Input: program $P \in HPP$, consisting of $m$ clauses $C_1, ..., C_m$ of the form $C_j = H_j : \mu_j \longleftarrow F_1^j : \nu_1 \wedge ... \wedge F_{r_j}^j : \nu_{r_j}$, formula $F = B_1 *_p ... *_p B_n$, and an interval $[a, b]$.
(1) $P_0 := \{H_j : \mu_j | body\ of\ C_j\ is\ empty\}$;
(2) FOR $i = 1$ TO $2m$ DO
(3)    guess $Fired(i)$;
(4)    FOR EACH $C_j \in Fired_P(i)$ DO
(5)       FOR $l = 1$ TO $r_j$ DO
(6)          Call Ent-$HPP^0(P_i, F_l^j : \nu_l)$;
(7)       END_DO
(8)    END_DO
(9)    $P_i := P_{i-1} \cup \{H_j : \mu_j | C_j \in Fired(i)\}$
(10) END_DO
(11) Call Ent-$HPP^0(P_{2m}, F : [a, b])$
(12) Output "Yes" if all (nondeterministic) calls of Ent-$HPP^0$ were successful.

The following result establishes that algorithm **Ent-HPP** correctly computes entailment in nondeterministic polynomial time.

**Lemma 2** *Algorithm Ent-HPP determines nondeterministically if $P \models F : [a,b]$ in polynomial time.*

This result provides an upper bound in the following claim.

**Theorem 4** *The entailment problem is NP-complete for the classes $HPP$ and $HPP_{k,r}(k \geq 3)$.*

The proof of NP-hardness can be obtained, by reducing a well-known NP-complete problem **3-Dimensional Matching** to the entailment problem for the class $HPP_{3,0}$.

So far we have have shown that entailment problem is polynomial for hp-programs in $HPP_1$ and is NP-complete for hp-programs in $HPP_k$, $k \geq 3$. We now turn our attention to $HPP_2$. Here, our results are most interesting — it will turn out that for many different types of p-strategies, the entailment problem is polynomially solvable, though this does not appear to be the case for all p-strategies.

Recall that given a graph $G = (V, E)$, a *matching*[13] is a set $E' \subseteq E$ such that no two edges in $E'$ share a common vertex. A matching $E'$ is *maximal* iff every edge in $(E - E')$ shares a vertex with some edge in $E'$. If $V = 2m$, we say a matching $E$ is *complete* iff every vertex $v \in V$ is the endpoint of some edge $e_v \in E'$. It will turn out that entailment is *polynomial time equivalent* to the following *generalized matching problem* on general graphs.

**Generalized weighted matching problem**

*Given an edge-weighted, undirected graph $G = \langle V, E, w \rangle$ and a goal weight combination function $c$, find a complete matching for which the goal function on weights of selected edges is maximized (minimized).*

More formally, we define two classes of "yes-no" matching problems:
$GWM_{max}(c) = \{G = \langle V, E, w : E \rightarrow [0,1] \rangle, B \in [0,1] | |V| = 2m\}$ and there exists a complete matching $\{e_1, ..., e_m\} \subseteq E$ such that $c(w(e_1), ..., w(e_m)) \geq B\}$
and
$GWM_{min}(c) = \{G = \langle V, E, w : E \rightarrow [0,1] \rangle, B \in [0,1] | |V| = 2m\}$ and there exists a complete matching $\{e_1, ..., e_m\} \subseteq E$ such that $c(w(e_1), ..., w(e_m)) \leq B\}$.

The following result shows that entailment in HPPs is polynomial-time equivalent to the above generalized matching problems.

**Theorem 5**
*(1) Let $P \in HPP_2$ use probabilistic strategies with combination functions $c = \langle c^1, c^2 \rangle$ where $c^1 \in C^1, c^2 \in C^2$ for some sets of functions $C^1$ and $C^2$. Then the entailment problem for $P$ and annotated basic formula $F : \mu$ is polynomially reducible to the problems $GWM_{max}(c^1)$ and $GWM_{min}(c^2)$, where $c^1 \in C^1, c^2 \in C^2$.*
*(2) Any generalized weighted matching problem for goal functions, satisfying axioms (a)-(d) of Definition 1, is reducible in polynomial time to entailment problems for hp-programs of $HPP_{2,0}$.*

It is well-known [13] that weighted matching problem is solvable in polynomial time for the sum of edges weights. This allows to get effective algorithms for almost all of strategies considered in [2].

**Corollary 2** *The entailment problem for the class of $HPP_2$ programs over strategies $S = \{inc, igc, pcc, igd, pcd, ncd\}$ is solvable in polynomial time.*

The above result is interesting because it provides polynomial results for programs in $HPP_2$ for all but one composition strategies studied in [2]. This leads to an interesting open question.

**Open question.** Is there polynomial time computable composition function $c = (c^1, c^2)$ satisfying axioms (a)-(d) for which generalized matching problem $GWM_{min}(c^2)$ ($GWM_{max}(c^1)$) is NP-complete?

### 3.3 Complexity of the Consistency Problem

In this subsection, we establish the complexity of determining if an HPP is consistent, i.e. is there a hybrid formula function $h$ that satisfies all rules in $P$?

It is easy to see that even a simple $HPP_1$ program containing two simple facts, viz. $a : [0,0]$, $a : [1,1]$, is inconsistent. The complicated interactions between logic and probabilities can engender more devious inconsistencies in HPPs.

The following result tells us that to check if $P$ is consistent, if our language allows $n$ ground atoms, we need to create only all ground basic formulas $F$ containing all the $n$ atoms and check if $h_P(F) \neq \emptyset$ for them. If so, $P$ is guaranteed to be consistent.

**Lemma 3** Let $P \in HPP$ over set $S = \{\rho_1,\ldots,\rho_m\}$ of p-strategies and let $A = \{A_1,\ldots,A_n\}$ be all the atoms found in $P$. Then $P$ is consistent iff for all formulas $F_i$ of form $F_i = A_1 *_{\rho_i} \ldots *_{\rho_i} A_n$, $i = 1,\ldots,m$, $h_P(F_i) \neq \emptyset$.

As in the case of the Entailment Problem, we summarize our results in three cases — where programs are from $HPP_1$, from $HPP_2$ and from $HPP_3$ or larger.

**Theorem 6**
(1) Given a program $P \in HPP_1$ its consistency can be established in polynomial time.
(2) Inconsistency problem for $HPP_2$ is polynomially reducible to $GWM_{\min}$ and $GWM_{\max}$. So, for $HPP_2$ programs over the set of p-strategies $\{inc, igc, pcc, igd, pcd, ncd\}$ the consistency problem is solvable in polynomial time.
(3) Consistency problem for $HPP$ is co-NP-complete.

We present here only the *nondeterministic* algorithm **InCon** which checks if an *arbitrary* HPP is inconsistent.

Algorithm InCon.
Input. An arbitrary HPP $P$.
1. Guess the shortest "inconsistent" formula $F$.
2. Guess two partitions of $F$ into $G_1 *_\rho \ldots *_\rho G_m$ and $H_1 *_\rho \ldots *_\rho H_k$ two sets of numbers: $x_1,\ldots,x_m$ and $y_1,\ldots,y_k$, $0 \leq x_i, y_j \leq 1$ such that $c_\rho^1(x_1,\ldots,x_m) > c_\rho^2(y_1,\ldots y_k)$.
3. Using **Algorithm Ent-HPP** check that for all $i \in \{1,\ldots m\}$ $P \models G_i : [0, x_i]$ and all $j \in \{1,\ldots k\}$ $P \models H_j : [y_j, 1]$.
4. If all calls to **Algorithm Ent-HPP** of previous step are successful then output "Yes".

## 4 Proof Procedure

In this section, we present a sound and complete proof procedure for HPPs. The first proof procedure for probabilistic programs, introduced in [2] and [17] is based upon expanding the program $P$ to a larger set of clauses (a closure of the program) and then resolving queries against that set. Since this procedure is computationally inefficient, other tabulation based proof procedures have also been developed. Here, we present a Hilbert-style proof system for ground hp-programs which guarantees that all proofs are polynomially bounded in length ! This is an interesting and counterintuitive result — it says that (the answers to) all queries to HPPs have at least one polynomial explanation. Let us now define the axioms and inference rules of the **proof system** $HGR_P$.

**Definition 15** Let $P$ be a ground hp-program over set $S$ of p-strategies. We define the formal system $HGR_P$ as follows:
1. <u>Axioms</u> of $HGR_P$ are all expressions of the form: $\overline{A:[0,1]}$, where $A \in B_L$.
2. <u>Inference Rules</u>. There are 7 types of inference rule schemas in $HGR_P$. One type (<u>Program</u>) depends on the clauses of program $P$ while other 6 types of inference rule schemas are independent of clauses in $P$ but do depend on which p-strategies are in $S$.
• <u>Program</u>: Let $F : \mu \leftarrow G_1 : \mu_1,\ldots,G_k : \mu_k \in P$,

$$\frac{G_1 : \mu_1 \ \ldots \ G_k : \mu_k}{F : \mu}$$

*Note: rules corresponding to clauses with empty body ($k = 0$) are actually axioms.*
• <u>A-Composition</u>: Let $A_1, A_2 \in B_L$, $\rho \in S$

$$\frac{A_1 : \mu_1 \quad A_2 : \mu_2}{(A_1 *_\rho A_2) : c_\rho(\mu_1, \mu_2)}$$

• <u>F-Composition</u>: $A_1,\ldots A_k, B_1\ldots B_k \in B_L, \rho \in S$

$$\frac{(A_1 *_\rho \ldots *_\rho A_k) : \mu_1 \quad (B_1 *_\rho \ldots *_\rho B_l) : \mu_2}{(A_1 *_\rho \ldots *_\rho A_k *_{\bar{\rho}} B_1 *_\rho \ldots *_\rho B_l) : c_\rho(\mu_1, \mu_2)}$$

• <u>Decomposition (cut)</u>: Let $\rho \in S$ $\quad \dfrac{(F \oplus_\rho G):\mu}{F:md_\rho(\mu)}$

• <u>Clarification</u>: $\dfrac{F:\mu_1 \quad F:\mu_2}{F:\mu_1 \cap \mu_2}$

• <u>Exchange</u>: Let $A_1,\ldots,A_k \in B_L$, $\rho \in S$, and let $\overline{B_1,\ldots,B_k}$ be a permutation of $A_1,\ldots,A_k$

$$\frac{(A_1 *_\rho \ldots *_\rho A_k) : \mu}{(B_1 *_\rho \ldots *_\rho B_k) : \mu}$$

• <u>Interval Weakening</u>: $\dfrac{F:\mu \quad \mu \subseteq \mu_1}{F:\mu_1}$

3. A finite sequence $C_1 \ldots C_r$ of annotated formulas is called an $\underline{HGR_P\text{-derivation}}$ iff each formula $C_j = F_j : \mu_j$ can be deduced from zero (in the case of *axiom)*, *one, or more previous of* $C_1 \ldots C_{j-1}$ *by applying one of the* inference rules *to them. We call formula* $C_r$ the result *of the* $HGR_P$*-derivation.*

*4. An annotated formula* $C = F : \mu$ *is derivable in* $HGR_P$ *iff there exists such an* $HGR_P$*-derivation* $C_1, \ldots C_r$ *that* $C_r = C$. *We denote it by* $P \vdash_{HGR_P} C$, *or just by* $P \vdash C$ *in the absence of other inference systems.*

**Theorem 7 (soundness of** $HGR_P$**)** *Let* $P$ *be an* $hp - program$ *and let* $Q$ *be a an hp-formula. If* $P \vdash_{HGR_P} Q$, *then* $P \models Q$.

It is known that all "natural" proof systems for standard classical propositional logic have proofs of exponential size (see e.g. [23]). But this is not the fact in our proof system $HGR_P$. The following result states that $HGR_P$ is both a *complete inference system* and that the length of the proofs in $HGR_P$ is polynomially bounded.

**Theorem 8** *For any* $P \in HPP$ *and annotated formula* $F : \mu$ *if* $P \vdash F : \mu$ *then there exists such an* $HGR_P$*-derivation* $C_1, \ldots C_r$ *that* $C_r = F : \mu$ *and* $r \leq O(|P|^2 + |F : \mu|)$.

## 5 Conclusions

As described in the introduction, there are numerous kinds of dependencies that might exist between uncertain events. Probability theory mandates that the probability of a complex event be computed not only in terms of the probabilities of the primitive events involved, but also it should take into account, dependencies between the events involved. Hybrid Probabilistic Programs (HPPs) [2] represent one of the first frameworks that allow a logic program to explicitly encode a variety of different probability assumptions explicitly into the program, for use in inferencing. Most existing frameworks for uncertainty in logic programming [4, 5, 8, 9, 12, 14, 15, 17, 19, 21, 7] do not permit this. A few important initial attempts to incorporate different probabilistic strategies were made by Thone *et al.*[22], and Lakshmanan [12], which culminated in an extension of the relational algebra that accommodated different probabilistic strategies [10]. In this paper, we have made three contributions. First, we have developed algorithms to efficiently perform a variety of computations for hybrid probabilistic programs. Each of these algorithms is "tuned" to fit the class within which an HPP falls (i.e. class $HPP_1$, $HPP_2$ or $HPP_r$, $r \geq 3$). We have given algorithmic complexity analyses of these problems. To date, with the exception of the work by Kiessling's group [7, 22] and by Lukasiewicz [15], almost no work on bottom up algorithms for computing probabilistic logic programs exists. Our algorithms are the first to apply not only to HPPs, but to have finer complexity bounds for different classes of HPPs.

Second, we have studied the computational complexity of the Entailment and Consistency problems for the abovementioned classes of HPPs. The results may be neatly summarized via the following table.

Table 1: Complexity results

| Problem | $HPP_1$ | $HPP_2$ | $HPP_r$ $r \geq 3$ |
|---|---|---|---|
| Entail-ment | poly-nomial | polynomial for $c_\rho \in \{inc, igc, pcc, igd, pcd, ncd\}$ | NP-complete |
| Consis-tency | poly-nomial | polynomial for $c_\rho \in \{inc, igc, pcc, igd, pcd, ncd\}$ | co-NP-complete |

In effect, this result says that from the point of view of complexity, it is possible to safely write HPPs over class $HPP_1$ (with any set of composition strategies), or over class $HPP_2$ (but with certain composition strategies only), and be guaranteed a polynomial computation. To our knowledge, this paper is the first paper to contain a detailed analysis of complexity results in probabilistic logic programs, though [10] contains some results for probabilistic relational algebra, and [12] contains some results for a different probabilistic framework.

Finally, we have described a proof system for HPPs that guarantees that for every $F : \mu$ that is a ground logical consequence of an HPP $P$, we have a polynomially bounded proof of $F : \mu$, which in turn, means that an *explanation* for $F : \mu$ is polynomially bounded. Though many proof systems have been developed for annotated logic programs they do not apply to probabilistic logic programs. Our proof system $HGR_P$ is new (and is also different from the proof system in [2]), and to our knowledge none of the existing proof systems for annotated logic have been shown to have polynomially bounded proofs (and hence succinct explanations).

### Acknowledgements.

The work of the first author had been partially supported by Russian Fundamental Studies Foundation (Grants 97-01-00973). The other authors were supported by the Army Research Office under Grants DAAH-04-95-10174, DAAH-04-96-10297, and DAAH04-96-1-0398, by the Army Research Laboratory under contract number DAAL01-97-K0135 and under Cooperative Agreement DAAL01-96-2-0002 Federated Laboratory ATIRP Consortium, by an NSF Young Investigator award IRI-93-57756, and by a

TASC/DARPA grant J09301S98061.